\documentclass[10pt,twocolumn,letterpaper]{article}

\usepackage{cvf}
\usepackage{times}
\usepackage{bm}
\usepackage{epsfig}
\usepackage{graphicx}
\usepackage{amsmath}
\usepackage{amssymb}
\usepackage{multirow}
\usepackage{booktabs}
\usepackage{colortbl}
\usepackage{xcolor}
\usepackage{array}
\usepackage{pifont}
\usepackage[T1]{fontenc}
\usepackage[utf8]{inputenc}
\usepackage{authblk}
\usepackage[pagebackref=true,breaklinks=true,letterpaper=true,colorlinks,bookmarks=false]{hyperref}
\usepackage{balance}

\cvffinalcopy 

\def\cvfPaperID{****} 

\definecolor{mycolor}{RGB}{241,240,255}

\ifcvffinal\fi

\begin{document}

\title{Instance-aware Dynamic Prompt Tuning for Pre-trained Point Cloud Models}

\author[1 \thanks{These authors contributed equally to this work.}]{
    Yaohua Zha}
\author[1 $^*$]{
    Jinpeng Wang}
\author[2 \thanks{Corresponding author.}]{
    Tao Dai}
\author[3]{
    Bin Chen}
\author[1]{
    Zhi Wang}
\author[1,4]{
    Shu-Tao Xia}
\affil[1]{
    Tsinghua Shenzhen International Graduate School, Tsinghua University}
\affil[2]{
    College of Computer Science and Software Engineering, Shenzhen University}
\affil[3]{
    Harbin Institute of Technology, Shenzhen
}
\affil[4]{
    Research Center of Artificial Intelligence, Peng Cheng Laboratory}

\maketitle
\ifcvffinal\thispagestyle{empty}\fi

\begin{abstract}
Pre-trained point cloud models have found extensive applications in 3D understanding tasks like object classification and part segmentation. 
However, the prevailing strategy of full fine-tuning in downstream tasks leads to large per-task storage overhead for model parameters, which limits the efficiency when applying large-scale pre-trained models. 
Inspired by the recent success of visual prompt tuning (VPT), this paper attempts to explore prompt tuning on pre-trained point cloud models, to pursue an elegant balance between performance and parameter efficiency. 
We find while instance-agnostic static prompting, e.g. VPT, shows some efficacy in downstream transfer, it is vulnerable to the distribution diversity caused by various types of noises in real-world point cloud data. 
To conquer this limitation, we propose a novel Instance-aware Dynamic Prompt Tuning (IDPT) strategy for pre-trained point cloud models. 
The essence of IDPT is to develop a dynamic prompt generation module to perceive semantic prior features of each point cloud instance and generate adaptive prompt tokens to enhance the model's robustness. 
Notably, extensive experiments demonstrate that IDPT outperforms full fine-tuning in most tasks with a mere 7\% of the trainable parameters, providing a promising solution to parameter-efficient learning for pre-trained point cloud models. 
Code is available at \url{https://github.com/zyh16143998882/ICCV23-IDPT}. 

\end{abstract}

\section{Introduction}

With the rapid development of 3D scanning technology, point clouds, as irregular point sets that represent 3D geometry, have been widely used in various fields and tasks. Deep learning-based point cloud processing techniques  \cite{guo2021pct,li2018pointcnn,ma2022rethinking,qi2017pointnet,qi2017pointnet++,wang2019dynamic,ReCon} have drawn considerable attention as they can directly process raw point cloud data while preserving its rich information. As a classic deep learning paradigm, fine-tuning the foundation model pre-trained \cite{dong2022autoencoders,liu2022masked,pang2022masked,yu2022point,zhang2022point} on massive raw point clouds in specific downstream tasks has achieved state-of-the-art performance. However, this approach is storage-intensive, as it requires storing and deploying a separate copy of the backbone parameters for each task.

\begin{figure}[t]
\centering
\includegraphics[width=1.0\linewidth]{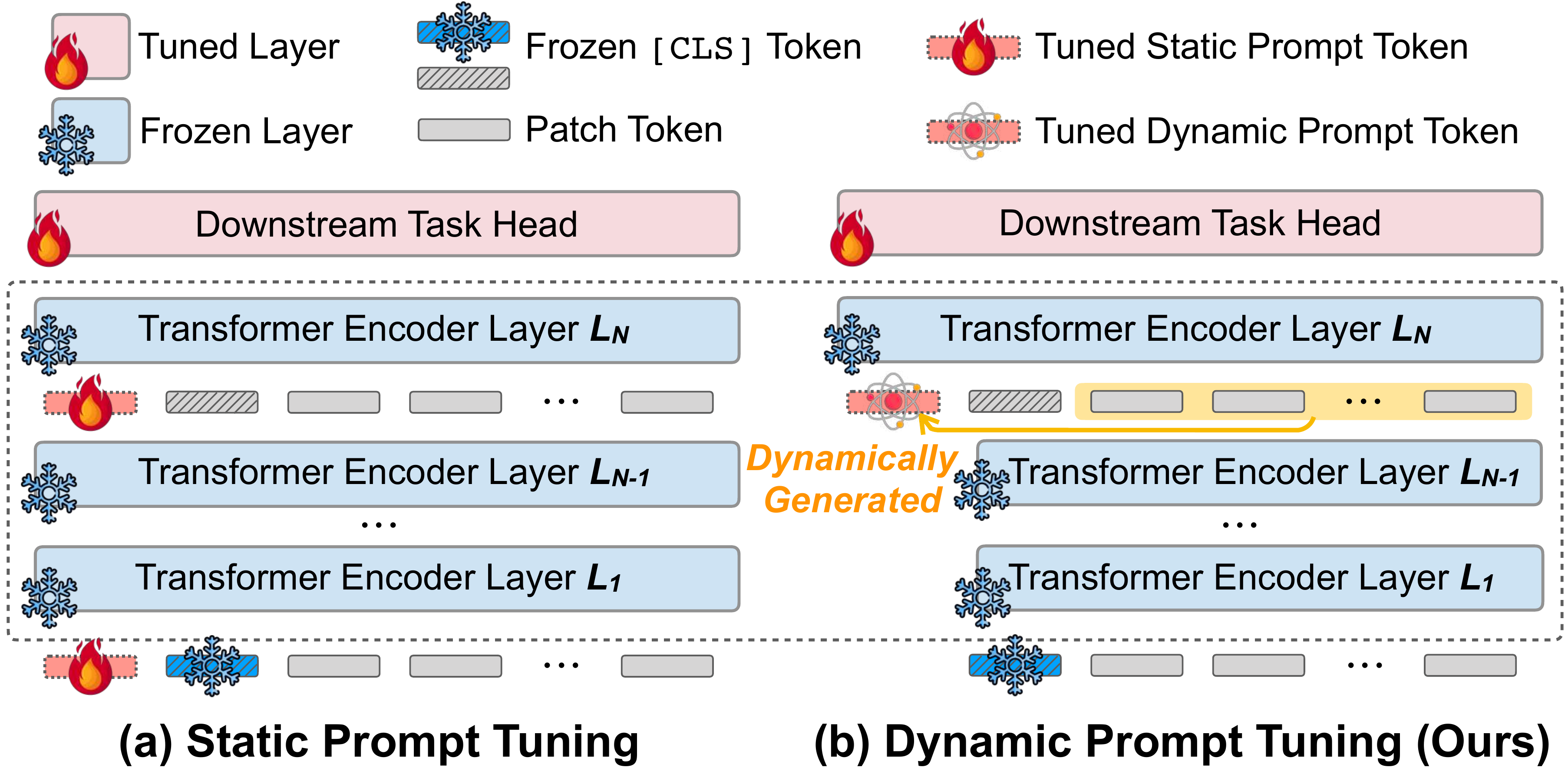} 
\caption{The pipeline of (a) the previous static prompt tuning in VPT \cite{jia2022visual} and (b) our dynamic prompt tuning. Unlike the static prompt tuning that is instance-agnostic, ours is adaptive to input by concatenating the instance-aware prompt generated by a prompt module into the last Transformer layer input.}
\label{intro}
\end{figure}

Recently, prompt tuning has surpassed fine-tuning in multiple downstream tasks in the language and image domains, significantly reducing storage requirements by fixing the parameters of a pre-trained model and introducing a small amount of task-specific learnable parameters into the input space. Although some work \cite{huang2022frozen,huang2022clip2point,wang2022p2p,zhang2022pointclip} have attempted  to introduce prompt into point cloud processing, they all relied on pre-trained image models \cite{radford2021learning,dosovitskiy2020image}.
To date, less research has been denoted to  prompt tuning in point cloud pre-trained models.

Inspired by the success of visual prompt tuning (VPT) \cite{jia2022visual}),  it is natural to adopt this idea to point clouds. 
As shown in Figure \ref{intro}(a), we call VPT a static prompting strategy because it introduces prompt tokens as a few learnable parameters concatenated to the input of a pre-trained point cloud model. 
The prompt tokens are shared by any input and therefore are instance-agnostic. 
Although such a strategy performs well on synthetic datasets (\eg ModelNet40 \cite{wu20153d}), it causes significant performance degradation on real scanned point cloud datasets (\eg ScanObjectNN \cite{uy2019revisiting}). Thus, static prompt tuning is not suitable for real point clouds,  where point clouds with different types of missing or noisy points belong to different distributions. These observations motivate us to  design a universal prompt-tuning strategy for both synthetic and real point clouds.

To address this issue, we proposed an Instance-aware Dynamic Prompt Tuning (IDPT) for point cloud pre-trained models. As shown in Figure \ref{intro}(b), IDPT develops a prompt generation module to perceive the semantic prior features of each point cloud instance and produces adaptive prompts for different inputs. 
The proposed IDPT enables an adjusting effect to mitigate the adverse noises in point cloud instances and thus can enhance the robustness of pre-trained models. 
We insert IDPT into the last Transformer layer for a more accurate representation of point clouds. 
Extensive experiments demonstrate the effectiveness of IDPT. 
Typically, IDPT yields competitive performance compared with full fine-tuning but just requires about 7\% of trainable parameters in downstream transfer.   

The main contributions can be summarized as follows:
\begin{itemize}
    \item  To our best knowledge, this is the first exploration of prompt tuning on pre-trained point cloud models. 
    We reveal that VPT, the static prompting strategy, suffers from the distributional diversity issue caused by various types of noises in real-world point cloud data. 
    \item To address the shortcoming, we propose an Instance-aware Dynamic Prompt Tuning (IDPT) as an effective solution. We develop a dynamic prompt generation module to capture semantic prior features of each point cloud instance and generate adaptive prompt tokens to mitigate the noises. 
    \item Extensive experiments on a variety of downstream tasks show the competitive performance of IDPT with full fine-tuning in most tasks while requiring much less tunable parameters, \eg about 7\%.
\end{itemize}

\begin{figure*}[t]
\centering
\includegraphics[width=\textwidth]{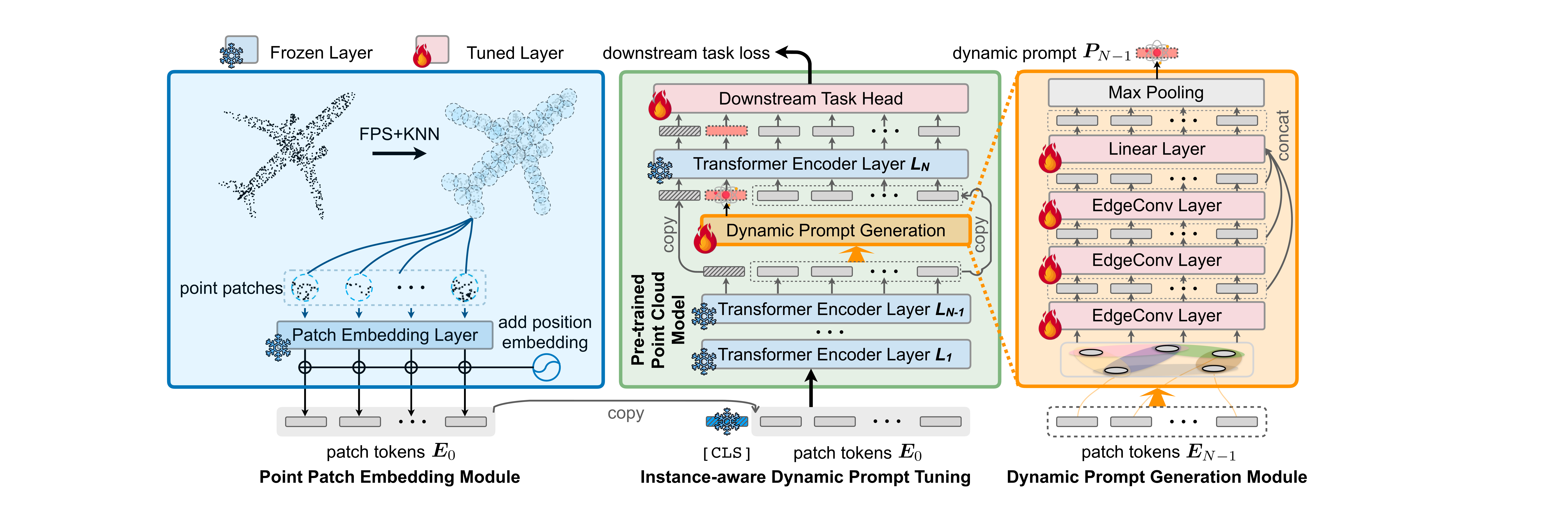} 
\caption{Overall pipeline of \textbf{I}nstance-aware \textbf{D}ynamic \textbf{P}rompt \textbf{T}uning (\textbf{IDPT}) for pre-trained point cloud models, which only updates the parameters of the dynamic prompt generation module and downstream task head during a downstream tuning task. 
To capture various sub-modes existing in the real-world data and enhance the robustness against noises (\eg, with different types of missing or noisy points), we design a dynamic prompt generation module with graph convolution \cite{wang2019dynamic} layers to aggregate multi-scale contextual features and dynamically generate instance-adaptive prompt. 
Empirically, inserting the dynamic prompt before the last transformer layer yields promising performance and enjoys decent efficiency at the same time. 
}
\label{framework}
\end{figure*}

\section{Related Work}

\subsection{Pre-training Point Cloud Models}

Recently, studies on pre-trained foundational models for 3D point clouds have achieved remarkable success. These approaches first apply a pretext task to pre-train the foundational model to learn the latent semantic information of the point cloud and then fine-tune the model 
 weights for the target task to achieve higher performance. Existing pre-train pretext tasks can be divided into discriminative tasks \cite{becker1992self,chen2020simple,gidaris2018unsupervised,wu2018unsupervised} and generative tasks \cite{baevski2022data2vec,dong2022autoencoders,he2022masked,lin2021end,liu2022masked,pang2022masked,yu2022point,zhang2022point}.
The discriminative approach distinguishes different views of the same point cloud instance from other instances, PointContrast \cite{xie2020pointcontrast} and CrossPoint \cite{afham2022crosspoint} explore the use of contrast learning of intra-domain and cross-domain features to obtain rich self-supervised information. Generation methods typically rely on an autoencoder to learn the latent features of the data by reconstructing the original input. Point-BERT \cite{yu2022point}, Point-MAE \cite{pang2022masked} and PointM2AE  \cite{zhang2022point}, based on masked autoencoders, have been very successful. Additionally, Point-DAE \cite{zhang2022point} explores a more general denoising autoencoder for point cloud learning by investigating more types of corruption beyond masking. ACT \cite{dong2022autoencoders} achieves a significant improvement on real scanned point clouds by using pre-trained language models and image models as cross-modal teachers to guide the learning of 3D self-supervised networks. However, the above methods all utilize full fine-tuning to adapt pre-trained models to various downstream tasks. Our work further explores how to reduce parameter storage in downstream tasks by utilizing prompt tuning, building upon the aforementioned approach.

\subsection{Prompt Learning in Computer Vision}

Prompt tuning involves adding specific prompt information to the input of a pre-trained model and adjusting downstream tasks to fit the pre-trained model. This is achieved by fixing the pre-trained model parameters and fine-tuning the prompt. It was first proposed in the language model \cite{brown2020language,gao2020making,lester2021power,liu2023pre,liu2021p,liu2021gpt} and gained popularity in the image model \cite{radford2021learning,rao2022denseclip,tsimpoukelli2021multimodal,zhou2022learning,zhou2022conditional} later due to its flexibility and high performance. CLIP \cite{radford2021learning} uses fixed class-specific text labels as prompts for prediction. Later, CoOp \cite{zhou2022learning} learns class-specific continuous prompts, and CoCoOp \cite{zhou2022conditional} builds upon CoOp by introducing a lightweight network to learn dynamic prompts for each instance. VPT \cite{jia2022visual} first introduces the continuous prompt tuning framework into image pre-trained models inspired by P-Tuning \cite{liu2021p}. Additionally, P2P \cite{wang2022p2p} achieved the first application of prompts in point clouds by learning color information in the input space of point cloud rendering images as prompts for the 2D backbone network. PointCLIP \cite{zhang2022pointclip} and CLIP2Point \cite{huang2022clip2point} project the point cloud as a depth map and then use the pre-trained CLIP \cite{radford2021learning} to understand the point cloud.  However, all the above work relies on pre-trained image models. Our work discusses the tuning of pre-trained point cloud models with appropriate prompting mechanisms.

\section{Methodology}
In this section, we first introduce the tuning pipeline for a pre-trained point cloud model (\S~\ref{subsec:prelim}). Next, we present the empirical observation of static prompt tuning (\eg VPT~\cite{jia2022visual}) and discuss its weaknesses (\S~\ref{subsec:obs}) that highlight our motivations. 
At last, we describe our Instance-aware Dynamic Prompt Tuning strategy in detail (\S~\ref{subsec:idpt}). 

\subsection{Preliminaries}
\label{subsec:prelim}
When fine-tuning a pre-trained point cloud model (\eg Point-MAE \cite{pang2022masked}), a point cloud $\bm X\in \mathbb{R}^{M\times 3}$ with $M$ points is first divided into $m$ point patches $\bm X'\in \mathbb{R}^{m \times k \times 3}$ via Farthest Point Sampling (FPS) and K-Nearest Neighborhood (KNN) algorithms, where each patch has $k$ local points. Then, all point patches will be embedded into a series of input tokens $\bm{E_0\in \mathbb{R}^{m\times d}}$ with positional encoding via a point patch embedding module. Next, we insert a classification token (\ie, \texttt{[CLS]}) $\bm c_{0}$ at the head of the patch embeddings and forward the token embeddings to the pre-trained model. Specifically, the forward process of each transformer layer is defined as
\begin{gather}
    \label{eq1}
    [\bm c_i; \bm E_i] = f_i([\bm c_{i-1}; \bm E_{i-1}]),\ i=1,2,\cdots,N,
\end{gather}
where $f_i$ denotes the $i$-th transformer encoder layer. $N$ is the total transformer layer number of the pre-trained backbone.
Finally, the model makes predictions by building a task-specific head $g_h$ upon the output of the pre-trained backbone: 
\begin{gather}
    \label{eq2}
    \bm y = g_h([\bm c_N; \bm E_N]).
\end{gather}
All the parameters of $\{f_i\}_{i=1}^N$ and $g_h$ will be updated in a downstream tuning task, which burdens the storage cost for per-task model weights.

Recently, prompt \cite{zhou2022learning, jia2022visual} has shown to be effective for parameter-efficient tuning. 
The basic idea of prompt tuning is to insert a few learnable prompt tokens into the input token sequence, \ie, we modify Eq.(\ref{eq1}) and Eq.(\ref{eq2}) by
\begin{gather}
    \label{eq3}
    [\bm c_i; \bm P_i; \bm E_i] = f_i([\bm c_{i-1}; \bm P_{i-1}; \bm E_{i-1}]),\ i=1,2,\cdots,N, \\
    \label{eq4}
    \bm y = g_h([\bm c_N; \bm P_N; \bm E_N]), 
\end{gather}
where $\bm P_i$ is the inserted prompt tokens at the $i$-th layer. 
During the tuning process, we freeze the parameters of $\{f_i\}_{i=1}^N$ and only update prompt $\{\bm P_i\}_{i=1}^N$ and task-specific head $g_h$, which can largely reduce per-task storage cost. 

\subsection{Observation and Discussion}
\label{subsec:obs}

\begin{table}[htbp]
  \centering
  \resizebox{\linewidth}{!}{
    \begin{tabular}{lccccc}
    \toprule
    \multirow{2}[4]{*}{\textbf{Tuning Strategy}} & \multirow{2}[4]{*}{\textbf{\#TP (M)}} & \multirow{2}[4]{*}{\textbf{ModelNet40}} & \multicolumn{3}{c}{\textbf{ScanObjectNN}} \\
\cmidrule{4-6}          &       &       & \textbf{OBJ\_BG} & \textbf{OBJ\_ONLY} & \textbf{PB\_T50\_RS} \\
    \midrule
    A.Only Head Tuning & 0.27  & 93.2  & 87.40 & 87.13 & 80.33 \\
    B.VPT-Shallow \cite{jia2022visual} & 0.28  & 93.4  & 87.61 & 89.04 & 80.99 \\
    C.VPT-Deep \cite{jia2022visual} & 0.36  & 93.6  & 89.98 & 90.19 & 83.96 \\
    \rowcolor{mycolor} D.\textbf{IDPT (Ours)} & 1.69  & 94.4  & 93.63 & 93.12 & 88.51 \\
    \midrule
    E.Full Fine-tuning & 22.10  & 93.8  & 92.94 & 92.08 & 88.41 \\
    \bottomrule
    \end{tabular}%
  }
  \caption{Classification accuracy (\%) for different tuning strategies is reported. All experiments were conducted based on a pre-trained Point-MAE \cite{pang2022masked} model, and a simple rotation augmentation in ACT \cite{dong2022autoencoders} was employed on the ScanObjectNN \cite{uy2019revisiting} dataset. '\#TP' denotes the number of trainable parameters.
  }
  \label{table_obs}%
\end{table}%

\begin{figure*}[t]
\centering
\includegraphics[width=1.0\textwidth]{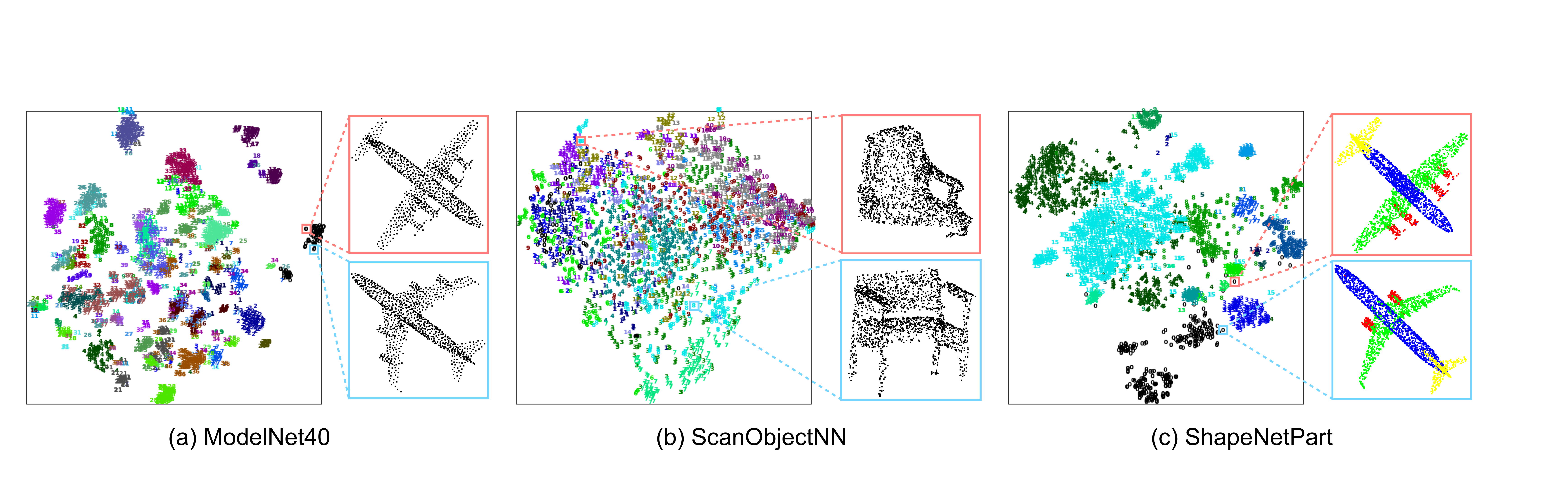} 
\caption{The t-SNE \cite{van2008visualizing} visualization of the point cloud features extracted from with the Point-MAE~\cite{pang2022masked} on three datasets: (a) ModelNet40, (b) ScanObjectNN, and (c) ShapeNetPart. Different from synthetic datasets (\eg ModelNet40 and ShapeNetPart) with clean and compact cluster structures, in real-world datasets (\eg ScanObjectNN), instances within the same category can present various sub-modes (\ie sub-clusters scattered in the embedding space) because real-world point clouds contain varying types of missing or noisy points. 
}
\label{raw_patch_tsne}
\end{figure*}

Inspired by the success of Visual Prompt Tuning (VPT) \cite{jia2022visual}, it is natural to extend such a prompting tuning strategy to point clouds, as shown in Figure~\ref{intro}(a). As the prompt tokens are instance-independent and shared by all samples during downstream tuning, we term this kind of strategy \emph{static prompt tuning}. Our empirical study showed that although VPT improves downstream performance compared with tuning the task head only, it underperforms fully fine-tuning by considerable margins. In particular, Table \ref{table_obs} presents the results of classification on several benchmark datasets. We can learn that static prompt tuning significantly reduces the number of trainable parameters (\ie, about 1\% to 2\% of backbone parameters) compared with fully fine-tuning. In terms of accuracy, although VPT-Deep performs well on synthetic datasets (\eg ModelNet40 \cite{wu20153d}), it presents significant performance degradation on real scanned point cloud datasets. For example, on the PB\_T50\_RS variant of ScanObjectNN \cite{uy2019revisiting} dataset, fully fine-tuning outperforms VPT-Deep by 4.5\% accuracy. 

Here we briefly discuss why static prompt tuning performs well on synthetic datasets but poorly on real scanned datasets. We adopt a perspective from Domain Adaptation (DA) \cite{aghajanyan2020intrinsic,ben2010theory,ben2006analysis,guo2022domain} and consider the transferring from pre-trained models to downstream tasks. Our goal is to bridge the source and target domains with different distributions so as 
 to enhance the prediction robustness of the pre-trained model. By definition, the source domain $p_s(\bm x_s)$ refers to the distribution of pre-training data, and the target domain $p_t(\bm x_t)$ refers to the distribution of downstream task data. Empirically, we found that 
\begin{equation}
  \label{eq5}
   p_t(\bm x_t) \ne  p_s(\bm x_s), 
\end{equation}
while we ask for a robust model such that 
\begin{equation}
  \label{eq6}
   p_t(\bm y_t|\bm x_t)=p_s(\bm y_s|\bm x_s),  
\end{equation}
where $\bm x$ denotes the input and $\bm y$ denotes the output. 
Therefore, the goal of domain adaptation is to find a transformation $\Phi(\cdot)$, such that 
\begin{equation}
  \label{eq7}
   p_t(\Phi(\bm x_t))=p_s(\Phi(\bm x_s)).
\end{equation}

Both fine-tuning and prompt tuning can be regarded as approaches to approximate the transformation $\Phi(\cdot)$. Fine-tuning adjusts all parameters of the pre-trained model to fit $\Phi(\cdot)$, whereas prompt tuning introduces additional prompt parameters to fit $\Phi(\cdot)$. From Table~\ref{table_obs}, we can learn that full fine-tuning achieves satisfactory performance by fitting all model parameters. In contrast, the static prompt shows inflexibility in mitigating the domain gap, suffering from the noises in the target domain.

We also provide an intuitive analysis of such a phenomenon. 
Specifically, Figure \ref{raw_patch_tsne} shows the t-SNE \cite{van2008visualizing} visualization of point cloud features extracted with Point-MAE \cite{pang2022masked} on the test sets of three datasets: (a) ModelNet40 \cite{wu20153d}, (b) ScanObjectNN \cite{uy2019revisiting}, and (c) ShapeNetPart \cite{yi2016scalable}).
It reflects the downstream task data distribution (\ie, target domain) to some degree. More details about the visualization of t-SNE are in the supplemental experiments.
As shown in the figure, on synthetic datasets like ModelNet40 and ShapeNetPart, instances from the same class tend to distribute in relatively clear and tight clusters. 
Quite differently on the real scanned dataset ScanObjectNN, instances from the same class scatter to many sub-clusters in the feature space, indicating the mixing of various sub-distributions corresponding to different sub-modes. 
Our intuition is that synthetic datasets like ModelNet40 and ShapeNetPart contain complete, uniform, and relatively clean point clouds, such as the airplanes shown in Figures \ref{raw_patch_tsne}(a) and \ref{raw_patch_tsne}(c). 
In contrast, ScanObjectNN consists of real scanned point clouds with varying types of missing or noisy points, making up different sub-modes within the same class. 
For example, Figure \ref{raw_patch_tsne}(b) shows two kinds of chairs: one is mostly missing, while the other is more complete. 
As static prompt tuning fails to capture various sub-modes in real-world data distribution, it highlights the necessity of an adaptive or dynamic prompting strategy.

\subsection{Instance-aware Dynamic Prompt Tuning}
\label{subsec:idpt}

To address the aforementioned issues of static prompt tuning, we propose \textbf{I}nstance-aware \textbf{D}ynamic \textbf{P}rompt \textbf{T}uning (\textbf{IDPT}). Figure \ref{framework} shows the pipeline of IDPT. 

\subsubsection{Dynamic Prompt Generation Module}
To capture various sub-modes existing in the real-world data and enhance the robustness against noises (\eg, with different types of missing or noisy points), we utilized EdgeConv~\cite{wang2019dynamic} at the patch level to perceive local point cloud shapes at a larger scale. 
Specifically, as shown in Figure \ref{framework}, point patch tokens $\bm E_{N-1}$ are processed by three EdgeConvs to generate three patch features at different scales. 
Then, the multi-scale patch features are concatenated and fed into a linear layer, followed by max pooling to generate an instance-aware dynamic prompt ${\bm P_{N-1}}$:
\begin{gather}
    \label{eq9}
    {\bm P_{N-1}} = \varphi_P(\bm E_{N-1}).
\end{gather}
$\varphi_P(\cdot)$ denotes the dynamic prompt generation module. 

Next, we forward ${\bm P_{N-1}}$ accompany with $\bm E_{N-1}$ to the last transformer layer $f_N$: 
\begin{gather}
    \label{eq10}
    [\bm c_N; {\bm P_N}; \bm E_N] = f_N([\bm c_{N-1}; {\bm P_{N-1}}; \bm E_{N-1}]).
\end{gather}

Finally, we concatenate \texttt{[CLS]} token $\bm c_{N}$, prompt token ${\bm P_{N}}$, and patch tokens $\bm E_{N}$ together before feeding them into a task head. 
The final prediction is made by:
\begin{gather}
    \label{eq11}
    \bm y = g_h([\bm c_N; {\bm P_N}; \bm E_N]).
\end{gather}

\subsubsection{Prompt Insert Position}
Perceiving various sub-modes in real-world point cloud data relies on high-level semantic information. 
As higher (or deeper) transformer layers grasp global semantic information (\eg density, shape, or categorical information) better, IDPT prefers to insert prompts at deeper transformer layers to ensure an accurate perception of point cloud semantics. In particular, we found that inserting the dynamic prompt before the last transformer layer yields robust performance and also enjoys decent efficiency. We provide detailed quantitative analysis in \S~\ref{subsubsec:prompt_insert}.

\section{Experiments}

We evaluated the performance of the proposed approach on classification, few-shot learning, and segmentation tasks. 
We used three widely used pre-trained models, Point-Bert \cite{yu2022point}, Point-MAE \cite{pang2022masked}, and ACT \cite{dong2022autoencoders}, as our baseline. 
Notably, our IDPT is a universal paradigm that can be applied to any pre-trained point cloud model. 

\subsection{Experiment Settings}

To ensure comparison fairness, we have used the same experimental settings as the default fine-tuning method for each baseline. This involves freezing the weights of the pre-trained point cloud model and only updating the parameters of the Prompt Model and Head during downstream task training. All experiments were conducted on a single GeForce RTX 3090 24GB. We have explored the performance of the simple rotation augmentation from ACT \cite{dong2022autoencoders} on the ScanObjectNN \cite{uy2019revisiting} dataset, which is denoted as \textcolor{red}{$\dagger$} in our table. 
In the downstream task experiments, we report the best result of 10 repeated experiments with different random seeds. In the ablation study, we report average results.

\subsection{Prompt Tuning in Downstream Tasks}
\subsubsection{Object Classification on Real-World Dataset} 
In the study of point cloud pre-training models, it is common practice to conduct pre-training on the ShapeNet \cite{chang2015shapenet} dataset, which typically only contains clean point clouds and assumes that all point clouds are identically distributed. However, in reality, point clouds often suffer from issues such as noise and missing points, resulting in a diverse distribution. We first assess our IDPT performance on the ScanObjectNN \cite{uy2019revisiting} dataset, which consists of about 15K point cloud samples by 15 categories. These objects are scanned indoor scene data, which are usually cluttered with background and occluded by other objects.

We conducted experiments on three variants of ScanObjectNN \cite{uy2019revisiting} (OBJ-BG, OBJ-ONLY, and PB-T50-RS). The results are shown in Table \ref{table1}, \textcolor{red}{$\dagger$} indicates that the pre-trained model used simple rotational augmentation of ACT during fine-tuning or prompt tuning, and without \textcolor{red}{$\dagger$} indicates the default augmentation method. We observed that: (i) We achieved state-of-the-art (SOTA) performance with IDPT on Point-MAE\textcolor{red}{$^\dagger$}. 
In comparison to the current state-of-the-art method ACT, we have achieved gains of 0.34\%, 1.21\%, and 0.3\% respectively in the three variants of ScanObjectNN, while utilizing only 7\% of its trainable parameters. 
(ii) Our IDPT outperforms full fine-tuning in most cases with fewer trainable parameters. These results demonstrate the excellent performance of our method on real scanned point clouds with various data distributions. We believe this is due to the introduction of a semantic prior of real point cloud data on the one hand, and fewer trainable parameters to mitigate overfitting on the other.

\begin{table}[t]
  \centering
  \resizebox{\linewidth}{!}{
    \begin{tabular}{lcccc}
    \toprule
    \textbf{Method} & \textbf{\#TP (M)} & \textbf{OBJ\_BG}\textcolor{blue}{($\uparrow$)} & \textbf{OBJ\_ONLY}\textcolor{blue}{($\uparrow$)} & \textbf{PB\_T50\_RS}\textcolor{blue}{($\uparrow$)} \\
    \midrule
    \multicolumn{5}{c}{\textit{Supervised Learning Only}} \\
    \midrule
    PointNet \cite{qi2017pointnet}  & 3.5   & 73.3  & 79.2  & 68.0 \\
    PointNet++ \cite{qi2017pointnet++}  & 1.5   & 82.3  & 84.3  & 77.9 \\
    DGCNN \cite{wang2019dynamic}  & 1.8   & 82.8  & 86.2  & 78.1 \\
    PointCNN \cite{li2018pointcnn} &  0.6  & 86.1  &  85.5 & 78.5 \\
    BGA-DGCNN \cite{uy2019revisiting} & 1.8   & -     & -     & 79.7 \\
    BGA-PN++ \cite{uy2019revisiting} &  1.5  & -     & -     & 80.2 \\
    DRNet \cite{qiu2021dense} & -     & -     & -     & 80.3 \\
    GBNet \cite{qiu2021geometric} & 8.8   & -     & -     & 80.5 \\
    SimpleView \cite{goyal2021revisiting} & -     & -     & -     & 80.8 \\
    PRANet \cite{cheng2021net} & 2.3   & -     & -     & 81.0 \\
    MVTN \cite{hamdi2021mvtn}  & -     & -     & -     & 82.8 \\
    PointMLP \cite{ma2022rethinking}  &       & -     & -     & 85.7 \\
    \midrule
    \multicolumn{5}{c}{\textit{with Self-Supervised Representation Learning (Full Fine-tuning)}} \\
    \midrule
    Transformer \cite{yu2022point} & 22.1  & 79.86 & 80.55 &  77.24 \\
    OcCo \cite{yu2022point}  & 22.1  & 84.85 & 85.54 & 78.79 \\
    Point-BERT \cite{yu2022point}  & 22.1  & 87.43 & 88.12 &  83.07 \\
    MaskPoint \cite{liu2022masked}  & 22.1  & 89.70 & 89.30 &  84.60 \\
    Point-MAE \cite{pang2022masked}  & 22.1  & 90.02 & 88.29 & 85.18 \\
    Point-M2AE \cite{m2ae}  & -  & 91.22 & 88.81 & 86.43 \\
    ACT\textcolor{red}{$^\dagger$} \cite{dong2022autoencoders}   & 22.1  & 93.29 & 91.91  & 88.21 \\
    Point-MAE\textcolor{red}{$^\dagger$} \cite{pang2022masked}  & 22.1  & 92.94 & 92.08 & 88.41 \\ 
    \midrule
    \multicolumn{5}{c}{\textit{with Self-Supervised Representation Learning (IDPT)}} \\
    \midrule
    \rowcolor{mycolor} \textbf{Point-BERT w/ IDPT} & \textbf{1.7} &  88.12 \textcolor{blue}{($\uparrow$ 0.69)}   & 88.30 \textcolor{blue}{($\uparrow$ 0.18)}      & 83.69 \textcolor{blue}{($\uparrow$ 0.62)} \\
    \rowcolor{mycolor} \textbf{Point-MAE w/ IDPT} & \textbf{1.7} & 91.22 \textcolor{blue}{($\uparrow$ 1.20)} & 90.02 \textcolor{blue}{($\uparrow$ 1.73)} & 84.94 \textcolor{gray}{($\downarrow$ 0.24)} \\
    \rowcolor{mycolor} \textbf{ACT\textcolor{red}{$^\dagger$} w/ IDPT} & \textbf{1.7}  & 93.12 \textcolor{gray}{($\downarrow$ 0.17)} & 92.26 \textcolor{blue}{($\uparrow$ 0.35)} & 87.65 \textcolor{gray}{($\downarrow$ 0.56)} \\ 
    \rowcolor{mycolor} \textbf{Point-MAE\textcolor{red}{$^\dagger$} w/ IDPT} & \textbf{1.7} & \textbf{93.63} \textcolor{blue}{($\uparrow$ 0.69)} & \textbf{93.12} \textcolor{blue}{($\uparrow$ 1.04)} & \textbf{88.51} \textcolor{blue}{($\uparrow$ 0.10)} \\ 
    \bottomrule
    \end{tabular}%
    }
  \caption{Classification results on three variants of ScanObjectNN dataset, and we report the number of trainable parameters (\#TP) and classification accuracy(\%). \textcolor{red}{$\dagger$} indicates that the pre-trained model used simple rotational augmentation of ACT \cite{dong2022autoencoders} during fine-tuning or prompt tuning. We achieve state-of-the-art performance with IDPT on Point-MAE\textcolor{red}{$^\dagger$} and our IDPT outperforms full fine-tuning in most cases with fewer trainable parameters.}
  \label{table1}%
\end{table}%

\subsubsection{Object Classification on Synthetic Dataset} 

\begin{table}[htbp]
  \centering
  \resizebox{\linewidth}{!}{
    \begin{tabular}{lcccc}
    \toprule
    \textbf{Method} & \textbf{ST?} & \textbf{\#TP (M)} & \textbf{Data Type} & \textbf{Accuracy (\%)} \\
    \midrule
    \multicolumn{5}{c}{\textit{Supervised Learning Only}} \\
    \midrule
    PointNet \cite{qi2017pointnet} & -     & 3.5   & 1k Points  & 89.2 \\
    PointNet++ \cite{qi2017pointnet++} & -     & 1.5   & 1k Points  & 90.7 \\
    DGCNN \cite{wang2019dynamic} & -     & 1.8   & 1k Points  & 92.9 \\
    PCT \cite{guo2021pct}   & N     & 2.9   & 1k Points  & 93.2 \\
    PVT \cite{zhang2021pvt}   & N     & -     & 1k Points  & 93.6 \\
    PointTransformer \cite{zhang2021pvt} & N     & -     & 1k Points  & 93.7 \\
    MVTN \cite{hamdi2021mvtn}  & -     & 11.2  & 12 Images  & 93.8 \\
    SimpleView \cite{goyal2021revisiting} & -     & -     & 6 Images   & 93.9 \\
    PointMLP \cite{ma2022rethinking} &       & 14.9  & 1k Points  & \textbf{94.5} \\
    \midrule
    \multicolumn{5}{c}{\textit{with Self-Supervised Representation Learning (Full Fine-tuning)}} \\
    \midrule
    Transformer \cite{yu2022point} & Y     & 22.1 & 1k Points  & 91.4 \\
    OcCo \cite{yu2022point}   & Y     & 22.1  & 1k Points  & 92.1 \\
    EPCL \cite{huang2022frozen}  & -     & -     & 1k Points  & 92.9 \\
    Point-BERT \cite{yu2022point}  & Y     & 22.1  & 1k Points  & 93.2 \\
    ACT \cite{dong2022autoencoders}   & Y     & 22.1  & 1k Points  & 93.7 \\
    Point-MAE \cite{pang2022masked} & Y     & 22.1  & 1k Points  & 93.8 \\
    MaskPoint \cite{liu2022masked} & Y     & 22.1  & 1k Points  & 93.8 \\
    Point-M2AE \cite{m2ae} & N     & 15.3     & 1k Points  & 94.0 \\
    CLIP2Point \cite{huang2022clip2point} & -     & -     & 10 Images  & 94.0 \\
    P2P \cite{wang2022p2p}   & -     & 1.2   & 1 Images   & 94.0 \\
    \midrule
    \multicolumn{5}{c}{\textit{with Self-Supervised Representation Learning (IDPT)}} \\
    \midrule
    \rowcolor{mycolor}\textbf{Point-BERT w/ IDPT} & Y     & \textbf{1.7} & 1k Points  & 93.4 \textcolor{blue}{($\uparrow$ 0.2)} \\
    \rowcolor{mycolor}\textbf{ACT w/ IDPT} & Y     & \textbf{1.7} & 1k Points  & 94.0 \textcolor{blue}{($\uparrow$ 0.3)} \\
    \rowcolor{mycolor}\textbf{Point-MAE w/ IDPT} & Y     & \textbf{1.7} & 1k Points  & \textbf{94.4} \textcolor{blue}{($\uparrow$ 0.6)} \\
    \bottomrule
    \end{tabular}%
  }
  \caption{Classification results on ModelNet40 \cite{wu20153d} dataset. ‘ST’ indicates whether the backbone is a standard Transformer \cite{vaswani2017attention} without any special design or inductive bias. ‘1k Points’ indicates that the input data is 1k points and ‘n Images’ indicates that the input data is n images. Our IDPT outperforms \textbf{full fine-tuning} in each baseline with fewer trainable parameters.}
  \label{table2}%
\end{table}%

We evaluate IDPT on the ModelNet40 \cite{wu20153d} dataset for object classification. ModelNet40 \cite{wu20153d} includes 12,311 clean 3D CAD models for 40 categories. Each point cloud is complete, uniform, and noise-free, and all point clouds in the dataset are independently and identically distributed. We follow standard protocols to split ModelNet40 into 9843 instances for the training set and 2468 for the testing set. Standard random scaling and random translation are applied for data augmentation during training. For fair comparisons, following previous work \cite{yu2022point,pang2022masked,dong2022autoencoders}, we also use the standard voting method \cite{liu2019relation} during testing. 

As shown in Table \ref{table2}, Point-MAE with IDPT achieves state-of-the-art performance with an accuracy of 94.4\%. This represents a 0.6\% improvement compared to fine-tuning. Additionally, other pre-trained models with IDPT, such as Point-BERT and ACT, demonstrate certain improvements compared to full fine-tuning. These results suggest that the incorporation of semantic priors of each instance can yield significant improvements.

\subsubsection{Few-shot Learning} 

\begin{table}[htbp]
  \centering
  \resizebox{\linewidth}{!}{
    \begin{tabular}{lcccc}
    \toprule
    & \multicolumn{2}{c}{5-way} & \multicolumn{2}{c}{10-way} \\
\cmidrule{2-5}          & 10-shot & 20-shot & 10-shot & 20-shot \\
    \midrule
    \multicolumn{5}{c}{\textit{with Self-Supervised Representation Learning (Full Fine-tuning)}} \\
    \midrule
    DGCNN-OcCo \cite{wang2021unsupervised}& 90.6±2.8 & 92.5±1.9 & 82.9±1.3 & 86.5±2.2 \\
    Transformer-OcCo \cite{yu2022point} & 94.0±3.6 & 95.9±2.3 & 89.4±5.1 & 92.4±4.6 \\
    Point-BERT \cite{yu2022point}  & 94.6±3.1 & 96.3±2.7 & 91.0±5.4 & 92.7±5.1 \\
    MaskPoint \cite{liu2022masked}  & 95.0±3.7 & 97.2±1.7 & 91.4±4.0 & 93.4±3.5 \\
    EPCL \cite{huang2022frozen}  & 95.1±2.7 & 97.3±1.6 & 91.1±4.2 & 93.5±3.8 \\
    Point-MAE \cite{pang2022masked} & 96.3±2.5 & 97.8±1.8 & 92.6±4.1 & 95.0±3.0 \\
    Point-M2AE \cite{m2ae} & 96.8±1.8 & \textbf{98.3±1.4} & 92.3±4.5 & 95.0±3.0 \\
    ACT \cite{dong2022autoencoders}  & 96.8±2.3 & 98.0±1.4 & \textbf{93.3±4.0} & \textbf{95.6±2.8} \\
    \midrule
    \multicolumn{5}{c}{\textit{with Self-Supervised Representation Learning (IDPT)}} \\
    \midrule
    \rowcolor{mycolor}\textbf{Point-BERT w/ IDPT}  & 96.0±1.7 \textcolor{blue}{ \textcolor{blue}{($\uparrow$)}} & 97.2±2.6 \textcolor{blue}{ \textcolor{blue}{($\uparrow$)}} & 91.9±4.4 \textcolor{blue}{ \textcolor{blue}{($\uparrow$)}} & 93.6±3.5 \textcolor{blue}{ \textcolor{blue}{($\uparrow$)}} \\
    \rowcolor{mycolor}\textbf{Point-MAE w/ IDPT} & \textbf{97.3±2.1} \textcolor{blue}{ \textcolor{blue}{($\uparrow$)}} & 97.9±1.1 \textcolor{blue}{ \textcolor{blue}{($\uparrow$)}} & 92.8±4.1 \textcolor{blue}{ \textcolor{blue}{($\uparrow$)}} & 95.4±2.9 \textcolor{blue}{ \textcolor{blue}{($\uparrow$)}} \\
    \rowcolor{mycolor}\textbf{ACT w/ IDPT} & 96.7±2.5 \textcolor{gray}{ \textcolor{gray}{($\downarrow$)}} & 98.2±0.9 \textcolor{blue}{ \textcolor{blue}{($\uparrow$)}} & 92.4±4.5 \textcolor{gray}{ \textcolor{gray}{($\downarrow$)}} & 95.5±3.0 \textcolor{gray}{ \textcolor{gray}{($\downarrow$)}} \\
    \bottomrule
    \end{tabular}%
  }
  \caption{Few-shot learning on ModelNet40. We report the average classification accuracy (\%) with the standard deviation (\%) of 10 independent experiments. Our IDPT achieved performance gains in most cases compared to full fine-tuning in few-shot learning.}
  \label{table3}%
\end{table}%

We conducted few-shot experiments on ModelNet40, using the n-way, m-shot setting, following previous works \cite{pang2022masked,yu2022point,zhang2022point}. The results for the settings of $n \in { 5, 10 }$ and $m \in {10, 20}$ are presented in Table \ref{table3}. Our IDPT achieved performance gains in most cases compared to full fine-tuning, demonstrating its efficacy in few-shot learning. 

\subsubsection{Part Segmentation}
\begin{table}[t]
  \centering
  \resizebox{\linewidth}{!}{
    \begin{tabular}{lccc}
    \toprule
    Methods & \#TP (M) &  $\mathrm{mIoU}_{c}$ & $\mathrm{mIoU}_{I}$ \\
    \midrule
    \multicolumn{4}{c}{\textit{Supervised Learning Only}} \\
    \midrule
    PointNet \cite{qi2017pointnet} & -  & 80.39 & 83.7 \\
    PointNet++  \cite{qi2017pointnet++} & -  & 81.85 & 85.1 \\
    DGCNN \cite{wang2019dynamic} & - & 82.33 & 85.2 \\
    \midrule
    \multicolumn{4}{c}{\textit{with Self-Supervised Representation Learning (Full Fine-tuning)}} \\
    \midrule
    Transformer \cite{yu2022point} & 27.09 & 83.42 & 85.1 \\
    OcCo \cite{yu2022point} & 27.09 & 83.42 & 85.1 \\
    MaskPoint \cite{liu2022masked} & - & 84.60 & 86.0 \\
    Point-BERT \cite{yu2022point} & 27.09 & 84.11 & 85.6 \\
    Point-MAE \cite{pang2022masked} & 27.06 & 84.19 & 86.1 \\ 
    ACT \cite{dong2022autoencoders}  & 27.06 & \textbf{84.66} & \textbf{86.1} \\
    \midrule
    \multicolumn{4}{c}{\textit{with Self-Supervised Representation Learning (Prompt Tuning)}} \\
    \midrule
    Point-MAE w/ VPT & 5.35 & 83.64  & 85.4  \\
    Point-MAE w/ IDPT & 5.69 & 83.79  & 85.7  \\
    ACT w/ VPT & 5.35 & 83.48 & 85.4 \\
    ACT w/ IDPT & 5.69 & 83.82  & 85.9 \\
    \bottomrule
    \end{tabular}%
    }
    \caption{Part segmentation results on the ShapeNetPart dataset. The mean IoU across all categories, \ie, $\mathrm{mIoU}_{c}$ (\%), and the mean IoU across all instances, \ie, $\mathrm{mIoU}_{I}$ (\%) are reported.}
  \label{part}%
\end{table}%

For part segmentation, we follow previous work \cite{dong2022autoencoders,pang2022masked} to add prompts to the input of 3-rd, 7-th and 11-th layers and the task head. Since we empirically observed that using a single-layer MLP achieves comparable performance to the three-layer EdgeConv architecture in the segmentation task, we adopt a simple single-layer MLP as the dynamic prompt generation module at each layer to reduce the number of trainable parameters. 

According to experimental results in Table \ref{part}, 
our IDPT outperforms the static prompting strategy, VPT \cite{jia2022visual}.
It verifies the effectiveness of our dynamic prompting strategy in part segmentation. 
But they are still inferior to full fine-tuning, We attribute the performance gap to the difficulty of fine-grained understanding of point clouds, which makes it challenging to  transfer pre-trained backbones with limited tunable parameters to the segmentation task. 
Fortunately, the design of instance-aware dynamics in IDPT helps to mitigate such a gap. 
We believe developing effective structure modeling mechanisms in the parameter-efficient tuning strategy is a promising direction for fine-grained point cloud tasks.

\subsection{Ablation Study}

To investigate the architecture design and tuning settings of our proposed strategy, we conducted extensive ablation studies on classification tasks in 2 variants of ScanObjectNN \cite{uy2019revisiting} (OBJ\_BG and OBJ\_ONLY).

\subsubsection{Compare IDPT with Other Tuning Strategies} 
In order to demonstrate the superiority of our proposed instance-aware dynamic prompt tuning over other tuning strategies (Head tuning, VPT \cite{jia2022visual} and full fine-tuning), we conducted extensive ablation studies as shown in Table \ref{table_obs}. The specific tuning strategies used are as follows: (A) Head tuning (where freezes the backbone and serves as a \textcolor{olive}{\emph{true}} reference for ``whether prompt tuning improves performance''), (B) VPT-Shallow \cite{jia2022visual}, (C) VPT-Deep \cite{jia2022visual}, (D) our IDPT, and (E) Fine-tuning.

Our IDPT tuning strategy (D) has outperformed other tuning strategies by achieving the highest level of performance while utilizing fewer trainable parameters. When compared to the baseline (A), we observed a \textbf{5.08\%}, \textbf{5.06\%}, and \textbf{8.18\%} improvement on the three variants of ScanObjectNN, respectively. Our strategy also significantly outperformed the traditional static prompt (B and C). Additionally, our method demonstrated a significant improvement over fine-tuning (E), as it greatly reduced the number of trainable parameters while still improving performance. Overall, these experiments clearly demonstrate that our IDPT tuning strategy is highly effective.

\subsubsection{Amount of Trainable Parameters and Structure}

\begin{table}[t]
  \centering
  \resizebox{\linewidth}{!}{
    \begin{tabular}{llccc}
    \toprule
    \textbf{Propmt Strategy} & \multicolumn{1}{c}{\textbf{Trainable Parameters Type}} & \textbf{Tr. Param.} & \textbf{OBJ\_BG} & \textbf{OBJ\_ONLY} \\
    \midrule
    w/o prompt & Head & 0.27 & 87.40 & 87.13\\
    \midrule
    VPT-Deep & 10 prompts + Head & 0.36  & 89.98 & 90.19 \\
    VPT-Deep & 28 prompts + Head & 0.52  & 90.02 & 90.53 \\
    VPT-Deep & 156 prompts + Head & 1.70  & 90.19 & 90.53 \\
    \midrule
    IDPT   & PM (1-layer MLP) + Head & 0.52  & 91.43 & 90.98 \\
    IDPT   & PM (3-layer MLPs) + Head& 1.49  & 91.64 & 91.34 \\
    IDPT   & PM (1 EdgeConv) + Head & 0.81  & 91.77 & 91.67 \\
    IDPT   & PM (2 EdgeConvs) + Head & 1.25  & 91.95 & 91.67 \\
    IDPT   & PM (3 EdgeConvs) + Head & 1.70  & \textbf{92.48} & \textbf{92.19} \\
    IDPT   & PM (1 Transformer layer) + Head & 2.14  & 92.03 & 91.22 \\
    \bottomrule
    \end{tabular}%
  }
  \caption{Effects of the number of trainable parameters and the structure of prompt generation module.}
  \label{table5}%
\end{table}%

We conducted experiments with varying numbers of trainable parameters to evaluate the effectiveness of our models. In VPT-Deep, we adjusted the trainable parameters by controlling the number of prompts in each layer input. Meanwhile, in IDPT, we experimented with various network structures, including MLP, graph convolution (EdgeConv), and Transformer layers. Our experimental results, as shown in Table \ref{table5}, demonstrate the following: (i) Using a single-layer MLP in IDPT resulted in significant improvements of 4.03\% and 3.85\% in OBJ\_BG and OBJ\_ONLY, respectively, when compared to the baseline without prompts. These results suggest that introducing semantic priors from the downstream task data can be highly effective. (ii) Increasing the number of parameters in VPT resulted in only limited performance improvements. It shows that merely increasing the number of parameters without introducing a semantic prior has limited performance improvement. (iii) For IDPT, EdgeConv proved to be an effective prompt module. Compared to MLP and Transformer, EdgeConv focuses more on local neighborhood information, allowing it to better perceive the specific shape of the point cloud and provide a better representation for downstream data semantic priors.

\subsubsection{Prompt Insert Position} 
\label{subsubsec:prompt_insert}

\begin{figure}[t]
\centering
\includegraphics[width=1.0\linewidth]{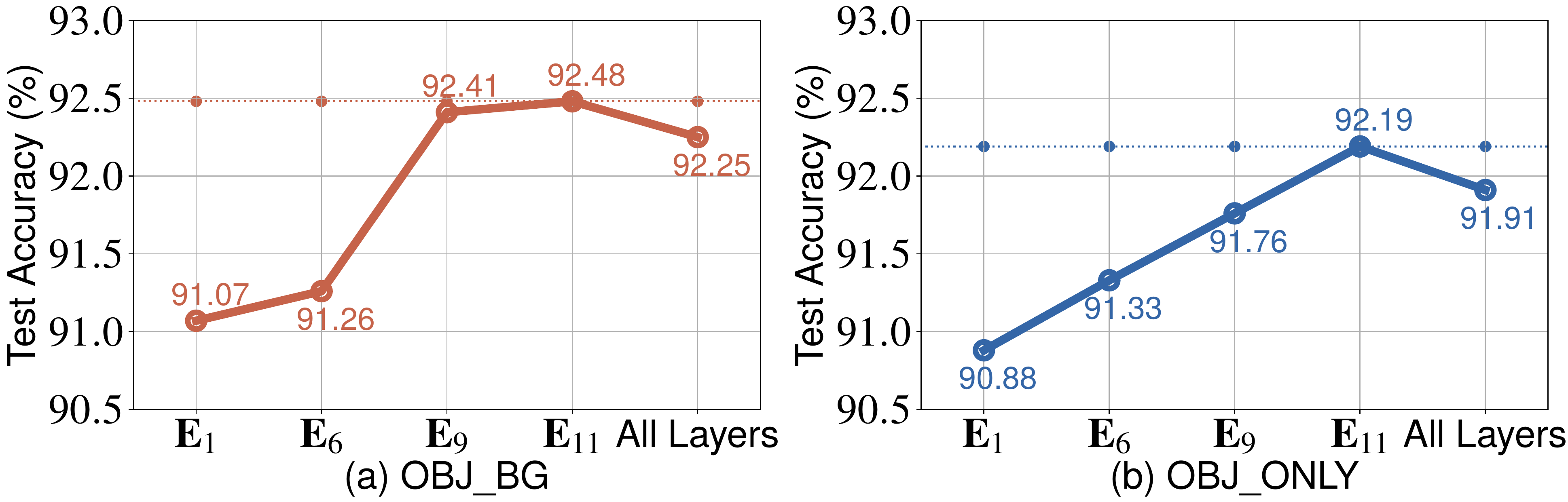} 
\caption{Effect of prompt insert position. $E_i$ means to insert our dynamic prompt into the input of $i$-th layer of the pre-trained Transformer. 'All Layers' means to insert our dynamic prompt in each layer of the Transformer using a prompt module with shared parameters.}
\label{abl_layer}
\end{figure}

We conducted an analysis to observe the impact of integrating our dynamic prompt in various depths of Transformer layers. The experimental results, as demonstrated in Figure \ref{abl_layer} (a) and (b), indicated that the addition of a prompt to deeper layers resulted in better performance. This can be attributed to the fact that our prompt generation module utilizes patch tokens to perceive the semantic priors of each point cloud, and deeper patch tokens comprise more comprehensive semantic information.

Additionally, we found that applying the prompt to all layers using a prompt module with shared parameters in the "all\_layer" setting did not produce satisfactory results. This is due to the fact that the shared prompt module can cause degradation in the semantic prior representation, and using independent parameters for each layer would result in an unacceptable increase in the number of parameters. As a result, we decided to add the prompt to the input of the final layer of the Transformer.

\subsubsection{Qualitative Analysis of the Ability to Approximate $\Phi(\cdot)$ in Downstream Adaptation} 

\begin{figure}[t]
\centering
\includegraphics[width=1.0\linewidth]{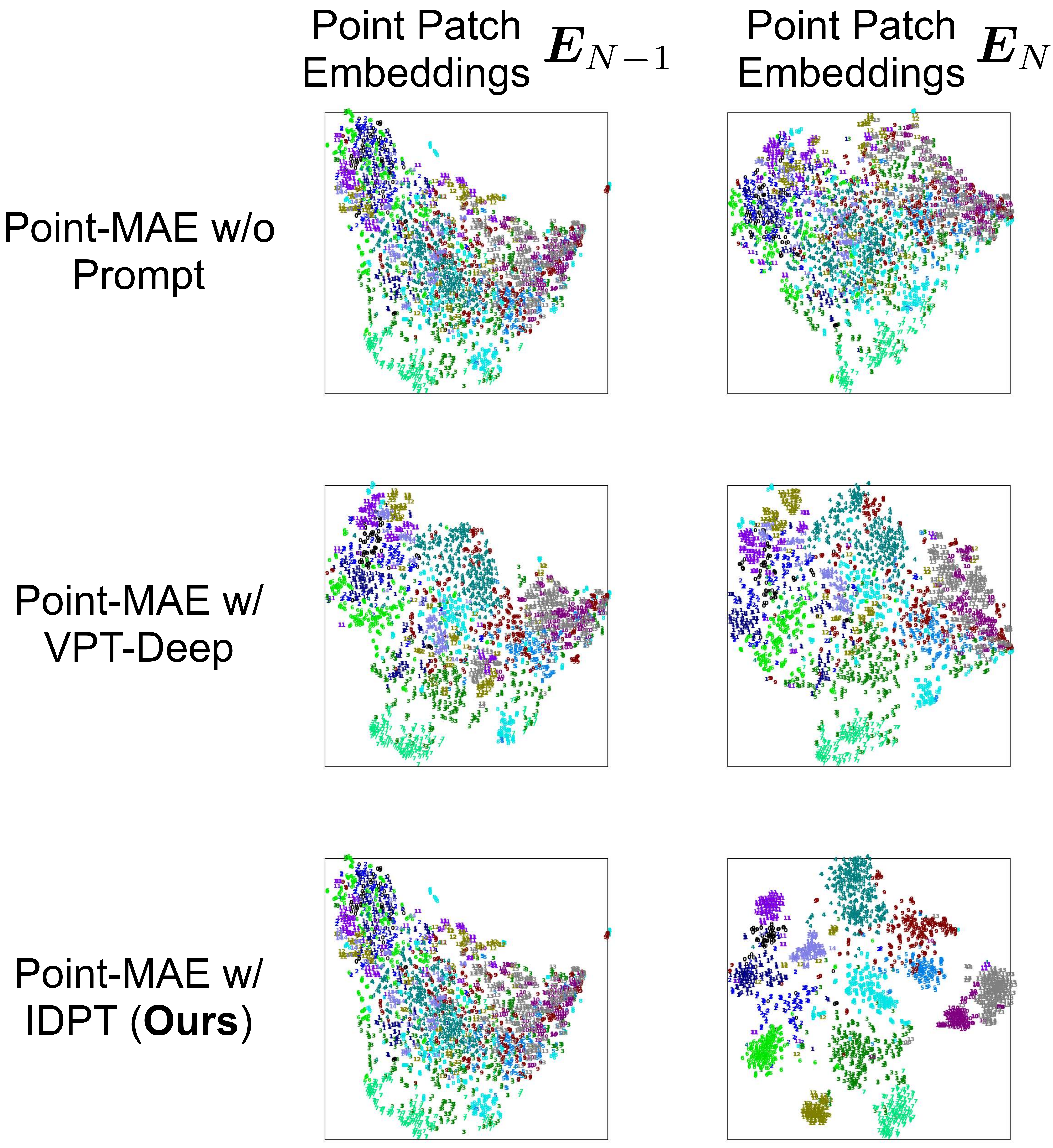} 
\caption{The t-SNE visualization of the point patch features extracted $\bm E_{N-1}$ and $\bm E_{N}$ from the test sets of ScanObjectNN (PB\_T50\_RS) using a pre-trained Point-MAE with different tuning strategies. This visualization partly reflects the approximation of the transformation function $\Phi(\cdot)$ by different tuning strategies.}
\label{tsne_prompt}
\end{figure}

We analyzed the effectiveness of various tuning strategies for the transformation function $\Phi(\cdot)$ by conducting a qualitative analysis of their fitting capability. The strategies we evaluated included (a) the pre-trained model, (b) VPT-Deep, and (c) IDPT. To visualize the input ($\bm E_{N-1}$) and output ($\bm E_{N}$) of all point patches in the $N$-th layer of the Transformer on ScanObjectNN, we utilized a pre-trained model based on Point-MAE. The results of our visualization are presented in Figure \ref{tsne_prompt}, which displays the visualization outcomes of the three tuning strategies.

The performance of different strategies for approximating the function $\Phi(\cdot)$ with the Transformer model varies. The pre-trained model (a) uses the Transformer with fixed parameters and shows the worst performance. VPT-Deep (b), which adds trainable static prompt parameters to all layers' inputs to approximate $\Phi(\cdot)$, resulting in a better input feature distribution $\bm E_{N-1}$ in the $N$-th layer than (a). However, the output features $\bm E_{N}$ are still scattered, indicating a poorer $\Phi(\cdot)$ approximate. IDPT takes a different approach by introducing the semantic prior of each instance and adding dynamic prompts to the input space of the $N$-th layer to approximate $\Phi(\cdot)$. Although the input features $\bm E_{N-1}$ at the $N$-th layer are as scattered as (a), the output features $\bm E_{N}$ are tightly clustered for the same category after concatenating $\bm E_{N-1}$ with our prompt through the same network as (a). It indicates that our strategy can effectively align the different distributions and is the best approximate strategy for $\Phi(\cdot)$.

\section{Conclusion}

In this paper, we investigate prompt tuning on pre-trained point cloud models to pursue the balance between performance and parameter efficiency. We found that the popular visual prompt tuning strategy cannot work well in real point clouds due to the distribution diversity. Therefore, we proposed instance-aware dynamic prompt tuning (IDPT) with a prompt generation mechanism to enhance the model's robustness in downstream transfer. Extensive experiments validated IDPT as a universal and effective solution.

\balance
{\small
\bibliographystyle{ieee_fullname}
\bibliography{egbib}
}

\clearpage
\appendix
\section{More Experimental Analysis}

\subsection{Number of Prompt Tokens}

In this section, we investigate the impact of prompt numbers in IDPT on classification tasks. By default, we use three layers of EdgeConv \cite{wang2019dynamic} and one layer of MLP to extract the semantic information $\bm{E_P\in \mathbb{R}^{m\times d}}$ from all patches and then use max pooling along the feature dimension to aggregate the semantic information of all patches to generate prompt $\bm{P_{N-1}\in \mathbb{R}^{1\times d}}$. 

To generate multiple representative prompts, we replace the max pooling operation along the feature dimension with a top-$K$ operation, resulting in $K$ prompts $\bm{P_{N-1}^{'}\in \mathbb{R}^{K\times d}}$. We then aggregate $\bm{P_{N-1}^{'}}$, $\bm c_{N-1}$, and $\bm E_{N-1}$ and feed them to the last transformer layer $\bm f_N$.
\begin{gather}
    \label{eq12}
    [\bm c_N; {\bm P_{N}^{'}}; \bm E_N] = f_N([\bm c_{N-1}; {\bm P_{N-1}^{'}}; \bm E_{N-1}]).
\end{gather}

For the classification head, we perform max pooling along the feature dimension of $\bm P_{N}^{'}$ to obtain $\bm P_{N}\in \mathbb{R}^{1\times d}$ as prompt-related input. 

\begin{figure}[h]
\centering
\includegraphics[width=\linewidth]{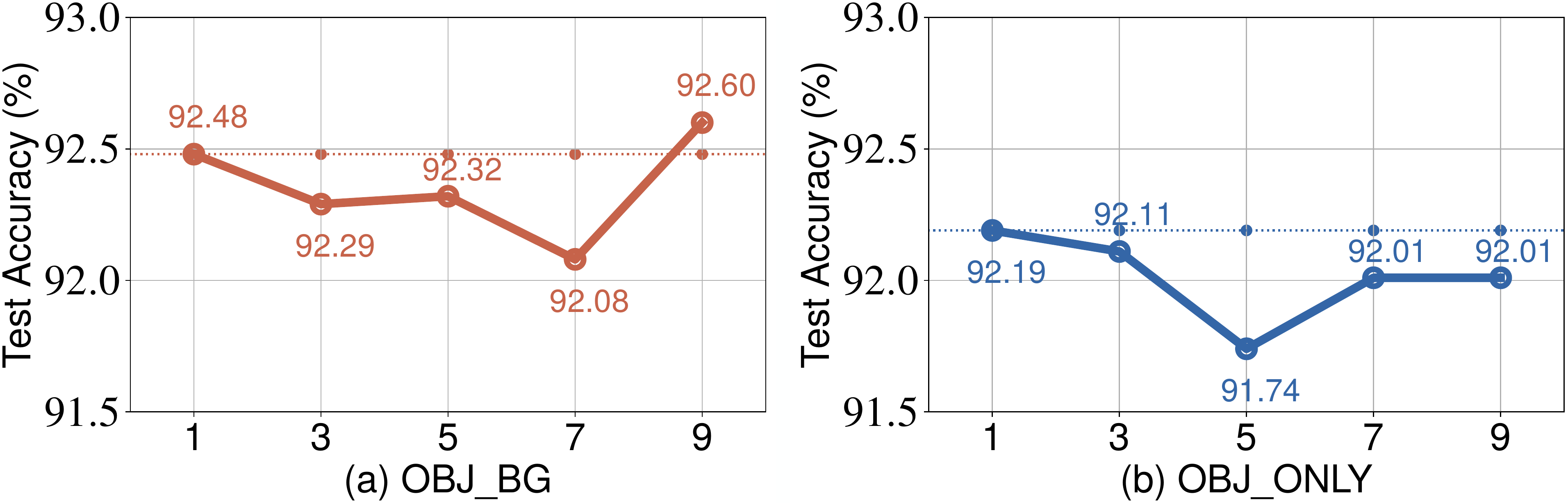} 
\caption{Effect of different numbers of prompts.}
\label{layer_num}
\end{figure}

We analyze the impact of different prompt numbers on classification tasks. Figure \ref{layer_num} presents the experimental results on two variants of ScanObjectNN. The results indicate that simply increasing the prompt number does not contribute to performance gain.  Therefore, we only set a single prompt in IDPT to improve efficiency.

\subsection{Insert Independent Prompt Generation Modules to All Layers}

In Figure \ref{abl_layer} of the main paper, we have demonstrated the effect of inserting prompts into multiple layers of the pre-trained point cloud model. 
Note that we share the prompt generation module among multiple layers in Figure \ref{abl_layer} in the spirit of parameter-efficient tuning. 
Nevertheless, it would be interesting to see the results of inserting independent prompt generation modules into different layers. 
In particular, here we provide the results of all-layer insertion, as shown in Table \ref{eachlayer}. 
The results indicate that incorporating a parameter-independent prompt generation module at every layer only brings marginal improvement with a significant increase of trainable parameters, deviating from the goal of parameter-efficient tuning. 
Regarding the empirical observations in Figure \ref{abl_layer} of the main paper and Table \ref{eachlayer}, we only insert the dynamic prompt generation module into the last layer of the pre-trained model.

\begin{table}[h]
  \centering
  \resizebox{\linewidth}{!}{
    \begin{tabular}{lccc}
    \toprule
    Trainable Parameters Type & \textbf{\#TP (M)} & \textbf{OBJ\_BG} & \textbf{OBJ\_ONLY} \\
    \midrule
    1  PM  + Head & 1.70  & 92.48 & 92.19 \\
    12  PM  + Head & 16.34 & \textbf{92.60} & \textbf{92.22} \\
    \bottomrule
    \end{tabular}%
  }
  \caption{Effect of inserting independent prompt modules to all layers. PM indicates the dynamic prompt generation module.}
  \label{eachlayer}%
\end{table}%

\subsection{Input of Downstream Task Head} 
\begin{figure}[h]
\centering
\includegraphics[width=1.0\linewidth]{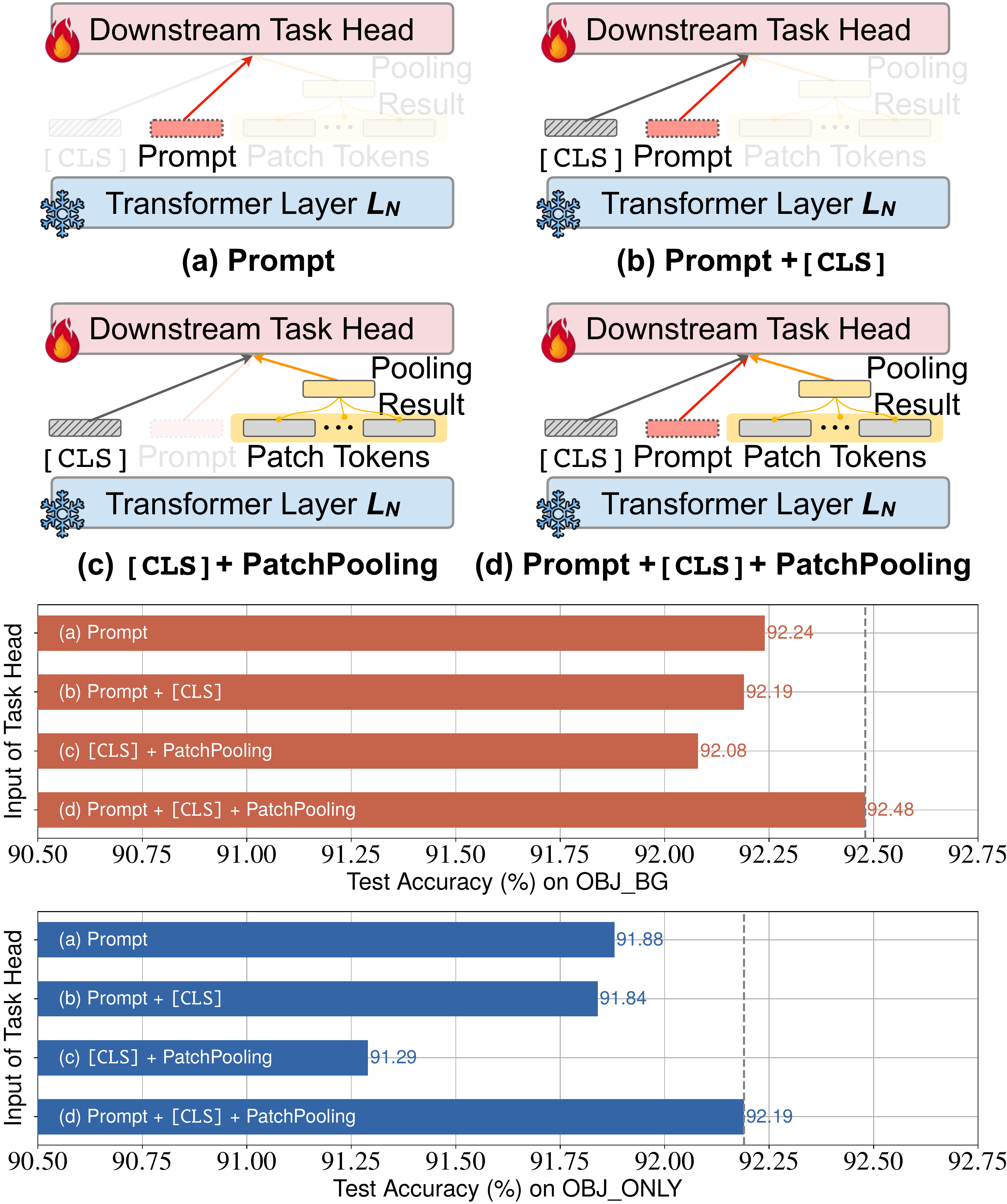} 
\caption{Effect of different head inputs.}
\label{abl_head}
\end{figure}

We conducted an investigation on the impact of the downstream task head's input features, which include the prompt token, the CLS token, and the max pooling of point patch tokens. The results are shown in Figure \ref{abl_head}. We found that the highest performance was achieved when all three features were included (d): prompt token, CLS, and point patch tokens. When using only the dynamic prompt token (b), the performance was still strong and only second to the previous case. However, removing our prompt token (c) resulted in a slight decline in performance. These results indicate that the dynamic prompt token plays a critical role in guiding the downstream task fitting, as it contains specific and semantic information about the task data.

Although omitting the prompt feature in the head results in a performance decline, there is still a significant improvement when compared to traditional static prompts, as shown in Table \ref{table5}. This suggests that our dynamic prompt is effective in aligning with different distributional data.

\subsection{Compare IDPT with Full Fine-tuning} 
\begin{figure}[h]
\centering
\includegraphics[width=0.95\linewidth]{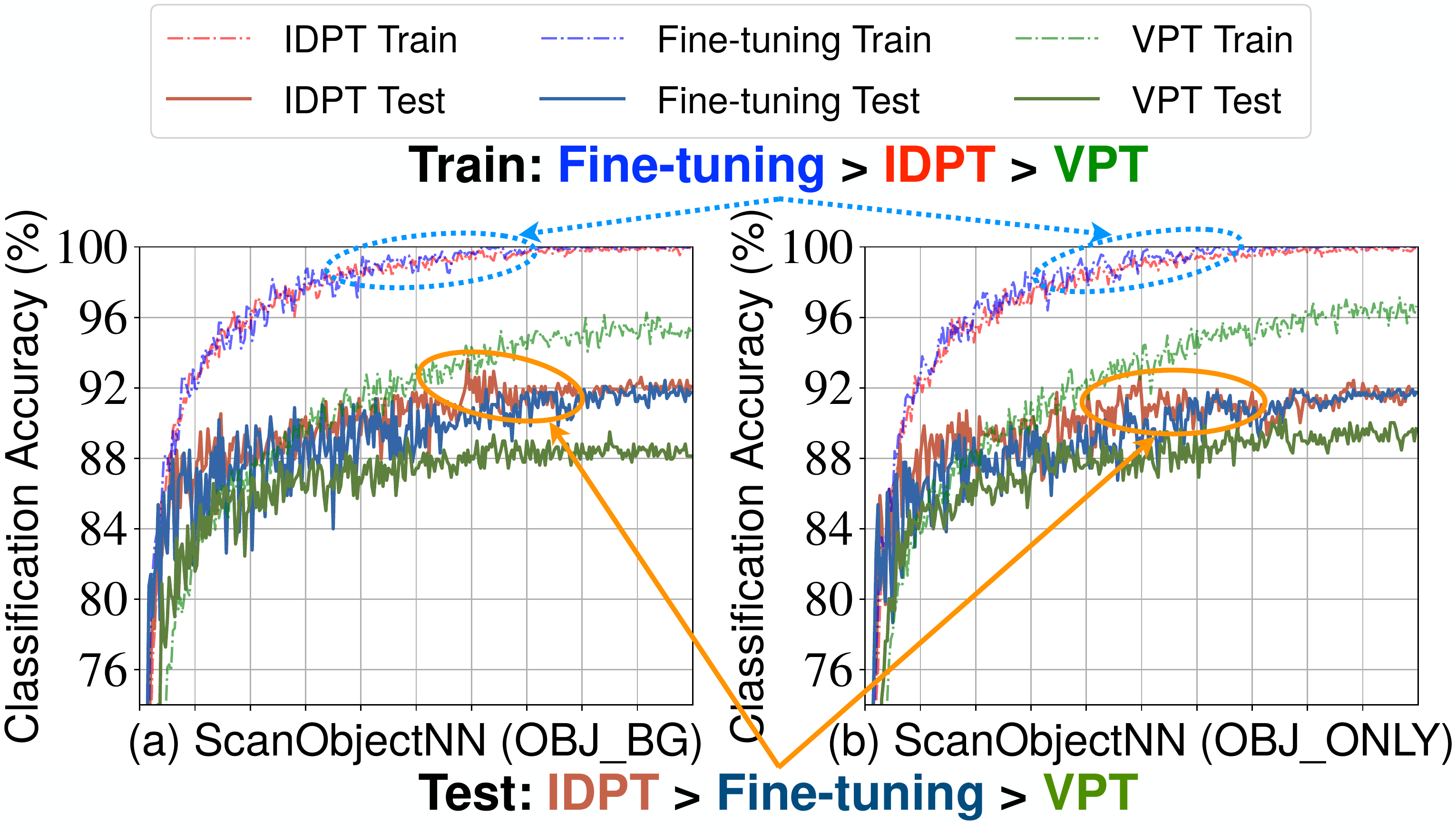} 
\caption{The convergence curves of training and testing.
}
\label{reason}
\end{figure}

Although fine-tuning with full trainable parameters enables more flexible adaptation, 
it is not always an optimal solution for downstream transfer. 
One possible issue is over-fitting the training set, which harms the generalizability. 
In particular, we investigate the accuracy dynamics of different tuning strategies with Point-MAE throughout training, as shown in Figure \ref{reason}. 
We can see IDPT outperforms fine-tuning on test sets, despite slightly inferior performance on train sets. 
This phenomenon demonstrates the structural flexibility of IDPT in domain adaptation and also indicates that it regularizes the model against over-fitting better. 
Besides, the efficacy of IDPT is not simply owed to the parameter-efficient setting, because VPT-Deep with a similar proportion of trainable parameters does not show satisfactory performance as IDPT does. 
Instead, the results provide evidence for the superiority of the proposed instance-aware dynamic prompt design.

\subsection{Convergence of Different Tuning Strategies}

\begin{figure*}[h]
\centering
\includegraphics[width=\textwidth]{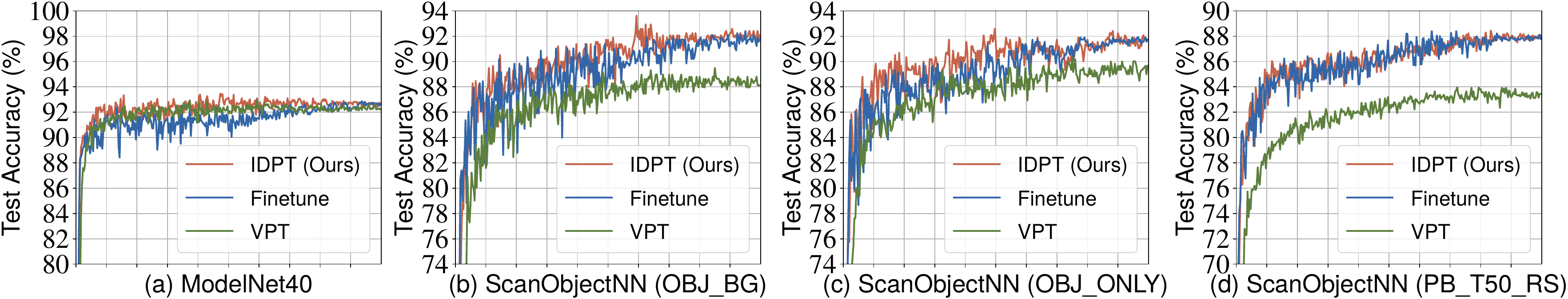} 
\caption{The classification accuracy curves of fine-tuning, VPT, and our IDPT strategy on two datasets.}
\label{curves}
\end{figure*}

In this section, we study how the performances of different tuning strategies change in the whole training process. 
The accuracy curves of fine-tuning, VPT, and IDPT on two datasets (\ie, ModelNet40 and ScanObjectNN) are illustrated in Figure \ref{curves}. 

As shown in Figure \ref{curves}, our IDPT strategy achieves significant improvements upon VPT. 
The performance of IDPT is competitive with fine-tuning on most datasets. 
Moreover, we can learn that IDPT yields faster convergence by incorporating prior semantic information of instances, revealing the merit of instance-aware dynamics for model adaptation.

\subsection{Demonstration of Sub-modes in Real-world Point Cloud Data}

\begin{figure*}[h]
\centering
\includegraphics[width=\textwidth]{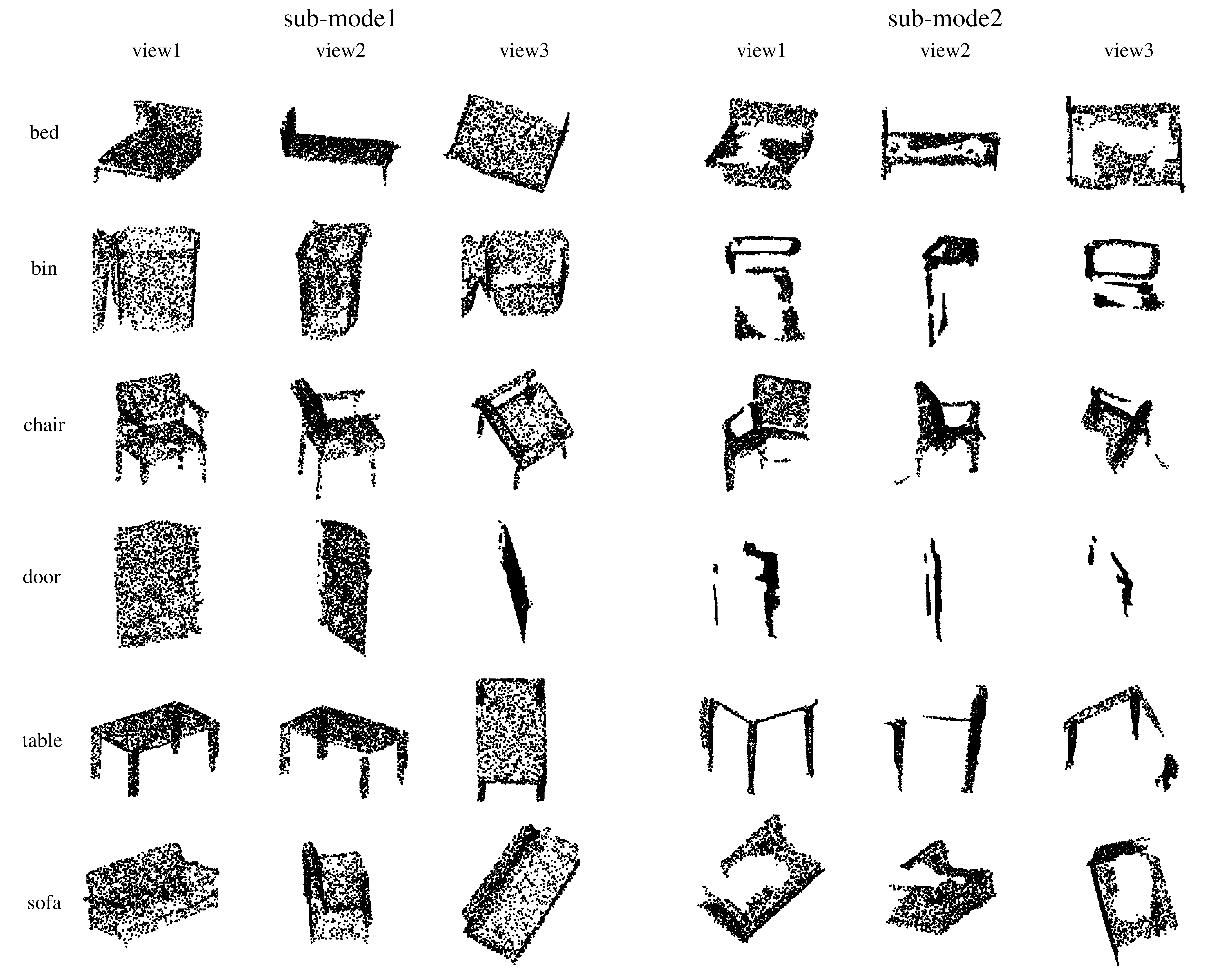} 
\caption{Different sub-modes in each category of ScanObjectNN}
\label{scan}
\end{figure*}

Due to the limitations of scanning techniques, 
it is prevailing to see various kinds of missing or noisy points in real-world point clouds, corresponding to different sub-modes in the data distribution. 
Such inconsistent noises will threaten the robustness of prompt-based adaptation, especially for static prompt strategies like VPT \cite{jia2022visual}.  
Here we would like to give some point cloud samples to facilitate an intuitive understanding about \emph{how different sub-modes look like}. 
Specifically, Figure \ref{scan} presents different missing types \wrt different categories in the ScanObjectNN dataset. 
We use sub\_mode1 and sub\_mode2 to differentiate missing types. 
For each scanned object, we show its projection images from three different viewpoints (\ie view1, view2, and view3) to simulate stereoscopy.

\end{document}